\pdfoutput=1
\documentclass[letterpaper]{article} 
\usepackage[preprint]{aaai2027}  
\usepackage[hyphens]{url}  
\usepackage{graphicx} 
\usepackage{natbib}  
\usepackage{caption} 
\usepackage{amsmath}
\usepackage{amssymb}
\usepackage{booktabs}
\usepackage{multirow}
\usepackage[table]{xcolor}
\usepackage{microtype}
\title{Face-D$^2$CL: Multi-Domain Synergistic Representation with Dual Continual Learning for Facial DeepFake Detection}
\author{
    Yushuo Zhang\textsuperscript{\rm 1},
    Yu Cheng\textsuperscript{\rm 1,\rm 2},
    Yongkang Hu\textsuperscript{\rm 1},\\
    Jiuan Zhou\textsuperscript{\rm 1},
    Jiawei Chen\textsuperscript{\rm 1},
    Zhaoxia Yin\textsuperscript{\rm 1}
}
\affiliations{
    \textsuperscript{\rm 1}East China Normal University\\
    Shanghai, China\\
    \textsuperscript{\rm 2}Shanghai Innovation Institute\\
    Shanghai, China
}

\begin{document}

\maketitle

\begin{abstract}
The rapid advancement of facial forgery techniques poses severe threats to public trust and information security, and imposes higher demands on the continual adaptation of DeepFake detection models. Although continual learning enables models to adapt to emerging forgery methods, existing approaches still face two key bottlenecks. On the one hand, they lack sufficient feature representation capacity for increasingly diverse and complex forgery traces. On the other hand, continual adaptation to new forgery distributions causes severe catastrophic forgetting of prior knowledge, substantially degrading detection performance. To address these issues, we propose Face-D\(^2\)CL, a framework for facial DeepFake detection. It leverages multi-domain synergistic representation to fuse spatial and frequency-domain features for the comprehensive capture of diverse forgery traces, and employs a dual continual learning mechanism that combines Real / Fake-aware Elastic Weight Consolidation (RF-EWC), which distinguishes parameter importance for real versus fake samples, and Domain-wise Orthogonal Gradient Constraint (D-OGC), which ensures updates to task-specific expert modules do not interfere with previously learned knowledge. This synergy enables the model to achieve a dynamic balance between robust anti-forgetting capabilities and agile adaptability to emerging facial forgery paradigms, all without relying on historical data replay. Extensive experiments demonstrate that our method surpasses current SOTA approaches in both stability and plasticity, achieving a 60.7\% relative reduction in average detection error rate. On unseen forgery domains, it further improves the average detection AUC by 7.9\% compared to the current SOTA method.
\end{abstract}

\section{Introduction}

Rapid DeepFake advances \cite{zhou2024migc} yield highly realistic forged faces, threatening information security, public trust, and personal privacy. Malicious uses such as identity impersonation and large-scale disinformation intensify these risks, while continually emerging forgery techniques further demand detectors with strong continual adaptation capability.

\begin{figure}[t]
    \centering
    \includegraphics[width=1.00\linewidth]{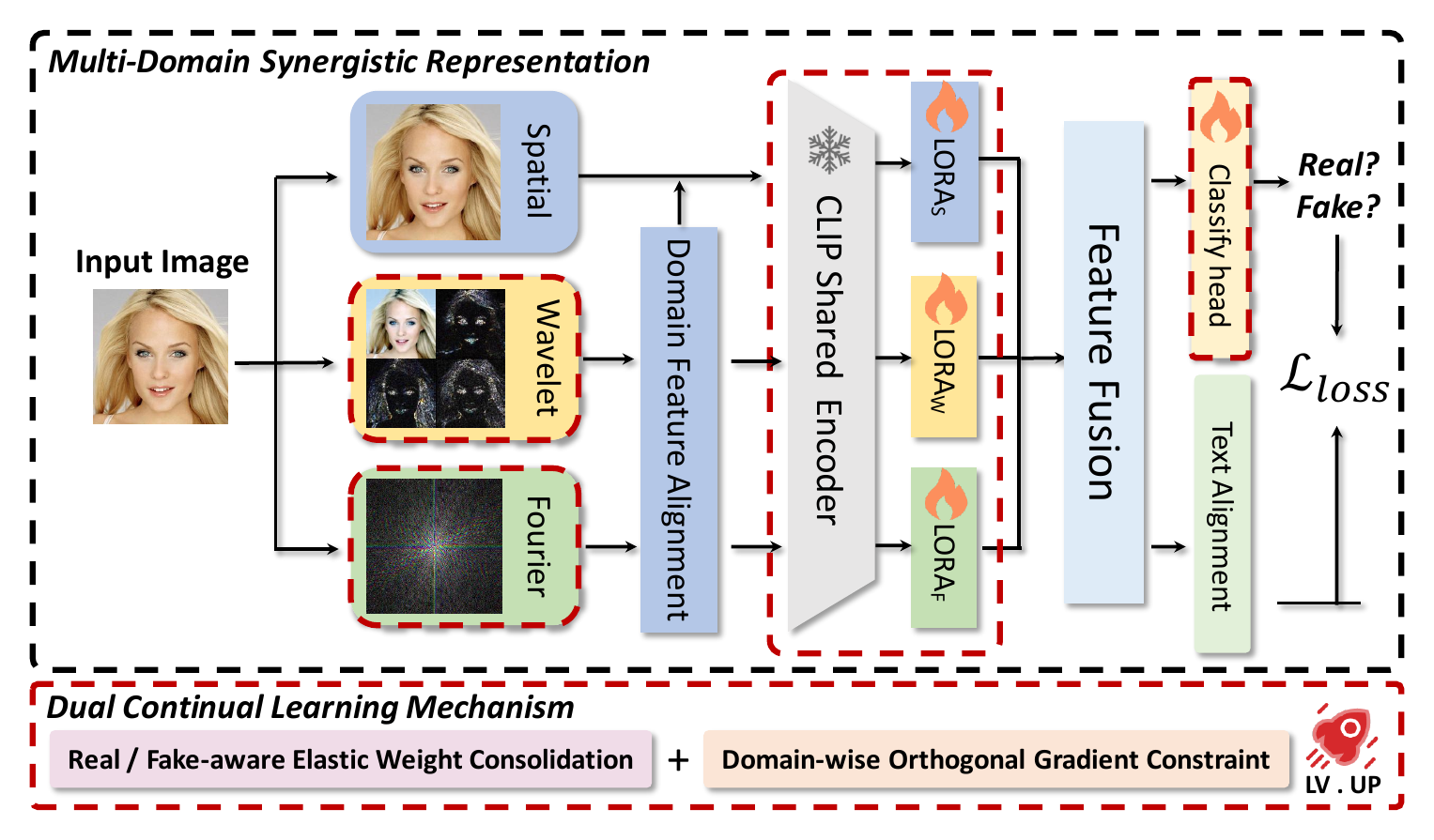}
    \caption{The pipeline of the proposed framework.}
    \label{fig:pipeline}
\end{figure}

Traditional face forgery detection methods typically learn discriminative features from fixed data and perform well on known forgery types \cite{chollet2017xception, tan2019efficientnet, qian2020thinking, zhangxinpeng_patchcraft, zhangxinpengtpami}. However, as generative models evolve rapidly from early GAN-based methods \cite{karras2018progressive, karras2021alias} to recent diffusion models \cite{rombach2022high, esser2024scaling, baldridge2024imagen} and commercial generators, forgery patterns have become increasingly diverse and complex, rendering static detectors inadequate for real-world use. A natural solution is to continually adapt detectors to newly emerging forgery data. Yet such adaptation often leads to catastrophic forgetting, where prior knowledge is disrupted or overwritten when learning new knowledge. The fundamental challenge lies in the distribution shift between emerging forgery domains and the original training distribution, hindering preservation of prior knowledge while adapting to new patterns.

Continual learning has emerged as a promising paradigm for face forgery detection, as it enables models to continually acquire new knowledge while retaining previously learned capabilities \cite{kirkpatrick2017overcoming, 8107520}. Existing continual learning methods mainly mitigate catastrophic forgetting through sample replay \cite{rebuffi2017icarl, chaudhry2019tiny, mai2021supervised}, parameter regularization \cite{kirkpatrick2017overcoming, 8107520}, and gradient projection strategies \cite{saha2021gradient, qiao2024prompt}. Building on these directions, several studies have investigated continual face forgery detection under different settings. CoReD \cite{kim2021cored} introduced distillation-based preservation to maintain previously learned knowledge, achieving promising results. DFIL \cite{pan2023dfil} combined center and hard-sample replay with multi-perspective distillation to alleviate catastrophic forgetting. SUR-LID \cite{cheng2025stacking} further maintained the global distribution of previous data through sparse uniform replay. More recently, SAIDO \cite{hu2026saido} proposed a scene-aware expert module and importance-guided gradient projection to achieve continual adaptation without relying on data replay. Despite these advances, existing methods still suffer from two key bottlenecks: insufficient feature representation for diverse forgery traces, which limits generalization to unseen forgery methods, and catastrophic forgetting during continual adaptation. This motivates the need for a more effective framework for continual face forgery detection.

To address these limitations, Face-D\(^2\)CL is introduced as a continual DeepFake detection framework that integrates multi-domain synergistic representation with a dual continual learning mechanism. Specifically, the multi-domain synergistic representation jointly extracts complementary forgery traces from three signal domains: the Spatial domain captures pixel-level inconsistencies and texture artifacts \cite{liu2021spatial, shiohara2022detecting}; the Wavelet domain reveals multi-scale reconstruction irregularities through frequency decomposition \cite{qian2020thinking}; and the Fourier domain exposes abnormal phase and magnitude patterns characteristic of generative models \cite{liu2021spatial}. By integrating these complementary cues, the proposed representation provides a more comprehensive view of forgery artifacts and enhances robustness across diverse forgery methods. For continual adaptation, a dual continual learning mechanism is further introduced. Replay raises privacy and storage concerns, standard EWC treats parameter importance uniformly, and orthogonal constraints alone cannot stabilize the fused multi-domain feature space. RF-EWC therefore uses separate real / fake Fisher importance for global stability, and D-OGC keeps domain-expert updates orthogonal to historical gradients for plasticity. Together they balance stability and plasticity. Experimental results show that Face-D\(^2\)CL achieves outstanding detection performance and continual learning capability across diverse face forgery methods, providing an efficient and practical solution for real-world face forgery detection. The main contributions of this work are summarized as follows:

\begin{itemize}\setlength{\itemsep}{0pt}\setlength{\parsep}{0pt}\setlength{\topsep}{0pt}\setlength{\partopsep}{0pt}
    \item A multi-domain synergistic representation module that extracts complementary forgery traces from the Spatial, Wavelet, and Fourier domains with a domain alignment strategy, providing broad feature coverage that enhances generalization across diverse methods.
    \item A dual continual learning mechanism that integrates RF-EWC and D-OGC in a complementary manner, where RF-EWC enhances stability by preserving important prior knowledge and D-OGC improves plasticity by enabling adaptation to emerging patterns while preserving prior domain-expert knowledge.
    \item Extensive experiments demonstrate that our method surpasses current SOTA approaches in both stability and plasticity, achieving a 60.7\% relative reduction in average detection error rate. On unseen forgery domains, it further improves the average detection AUC by 7.9\% compared to the SOTA method SAIDO.
\end{itemize}

\section{Related Work}

\subsection{Face Forgery Detection}

Early face forgery detection methods relied on handcrafted features to identify inconsistencies in head pose \cite{yang2019exposing}, eye blinking \cite{8630787}, or teeth details \cite{haliassos2021lips}. However, as forgery techniques advanced, these cues became less reliable. Recent deep networks learn discriminative features from large-scale datasets. Among them, spatial-domain methods such as Xception \cite{chollet2017xception} and EfficientNet \cite{tan2019efficientnet} achieve strong performance by capturing pixel-level artifacts. To improve generalization, frequency-domain analysis has been explored, with methods like F3-Net \cite{qian2020thinking} and SPSL \cite{liu2021spatial} mining high-frequency forgery traces. Another line uses data-level augmentation, such as Face X-ray \cite{li2020face} and Self-Blended Images (SBI) \cite{shiohara2022detecting}, which synthesize additional training data to expose forgery boundaries. More recently, contrastive learning learns domain-invariant representations \cite{sun2022dual, xu2022supervised}. Despite these advances, most methods assume static training and struggle to generalize to emerging forgery techniques, highlighting the need for continual learning.

\subsection{Continual Learning}


Continual learning enables models to acquire new knowledge sequentially while retaining prior information. Existing methods fall into three categories: regularization-based \cite{kirkpatrick2017overcoming, 8107520}, replay-based \cite{rebuffi2017icarl, chaudhry2019tiny, mai2021supervised}, and architecture-based methods \cite{rusu2016progressive, yan2021dynamically}. Regularization methods such as EWC and LwF constrain updates to important parameters but may overly limit adaptation to new tasks; replay methods such as iCaRL and Experience Replay preserve historical knowledge via past-sample subsets, but incur memory overhead and privacy concerns in sensitive settings; and architecture-based methods, such as Progressive Neural Networks and DER, mitigate forgetting via task-specific modules or dynamic capacity expansion, at higher complexity and parameter cost. Recently, gradient projection methods \cite{saha2021gradient, qiao2024prompt} show promise for mitigating forgetting without explicit replay.


In face forgery detection, CoReD \cite{kim2021cored} introduced distillation to retain prior-task knowledge, but may remain limited under substantial domain shifts. DFIL \cite{pan2023dfil} combined center / hard sample replay with multi-perspective distillation to alleviate forgetting. HDP \cite{sun2025continual} used adversarial perturbations as replay to enhance robustness against forgetting. SUR-LID \cite{cheng2025stacking} maintained global distribution via sparse uniform replay. However, replay adds memory overhead and privacy concerns from past-sample storage. SAIDO \cite{hu2026saido} proposed a scene-aware expert and importance-guided gradient projection for replay-free stability--plasticity balance, but its feature representation for diverse forgery traces remains insufficient, limiting unseen-forgery generalization and stable cross-task adaptation.

\section{Proposed Method}

To address insufficient multi-domain representation and catastrophic forgetting in continual DeepFake detection, we propose Face-D\(^2\)CL, unifying \textbf{Multi-Domain Synergistic Representation} with a \textbf{Dual Continual Learning Mechanism}. The former captures complementary forgery traces across Spatial, Wavelet, and Fourier domains; the latter stabilizes learning by jointly constraining global and low-rank parameters without data replay. Together they improve cross-domain features and continual learning, balancing plasticity and stability. For clarity, Table~\ref{tab:notation} summarizes the key notations used throughout this section.

\begin{table}[t]
\caption{Key notations.}
\label{tab:notation}
\centering
\small
\setlength{\tabcolsep}{3pt}
\resizebox{\columnwidth}{!}{%
\begin{tabular}{c p{0.7\columnwidth}}
\toprule
Symbol & Description \\
\midrule
$x_i$ & Input image \\
$f$ & Feature  \\
$\hat{f}$ & Normalized feature \\
$\theta$ & Model parameters \\
$g$ & Gradient \\
$\tilde{g}$ & Updated gradient after orthogonal projection \\
$\mu(\cdot), \sigma(\cdot)$ & Mean and standard deviation \\
$\mathcal{N}(\cdot)$ & Normalization operator \\
$\mathcal{L}$ & Loss \\
$\lambda$ & Balancing coefficient \\
$B$ & Batch size \\
\bottomrule
\end{tabular}
}
\end{table}

\subsection{Overall Framework}

\begin{figure*}[t]
\centering
\includegraphics[width=0.9\textwidth]{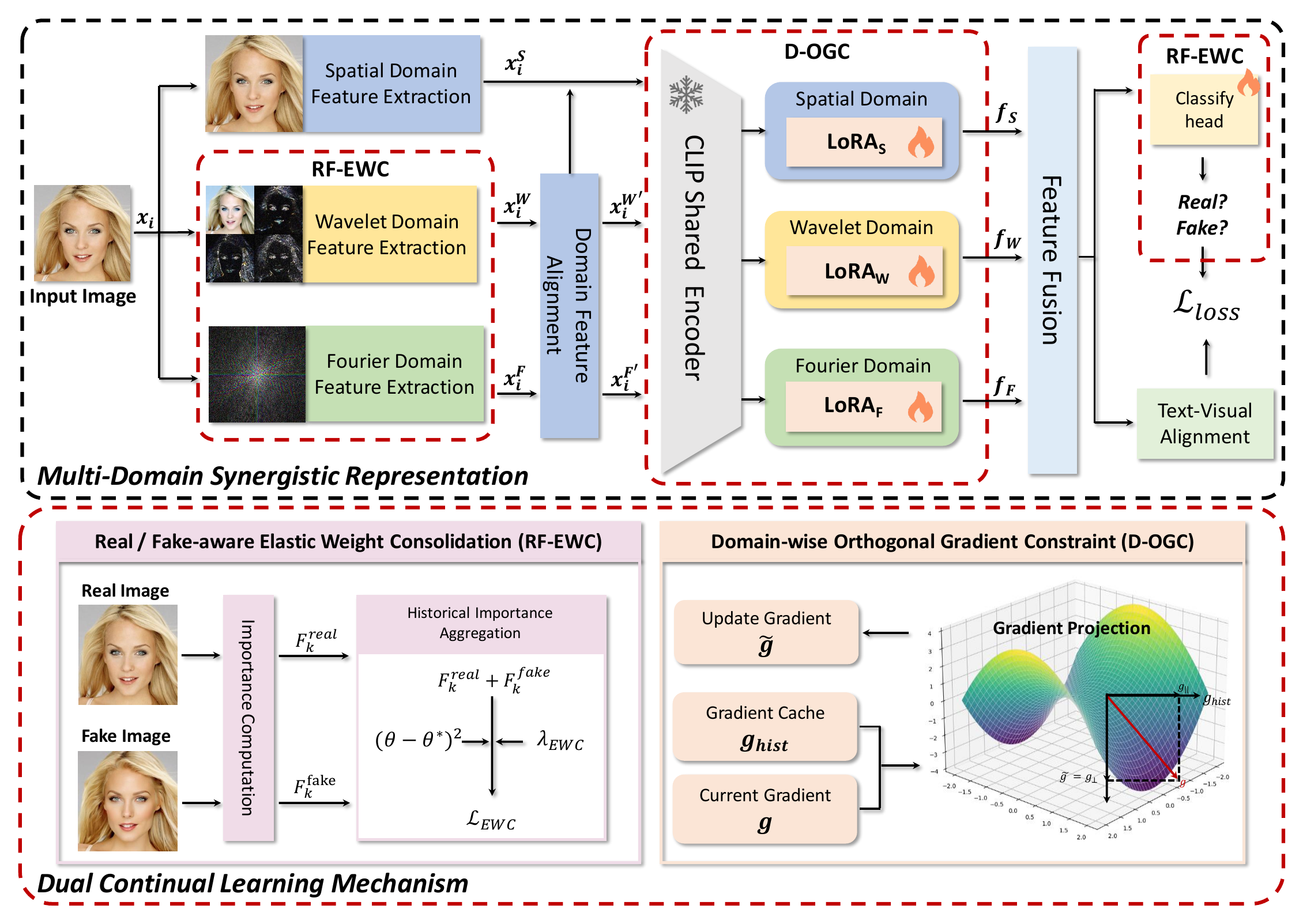}
\caption{Overall architecture of Face-D\(^2\)CL. The input face image is processed by three parallel branches (Spatial, Wavelet, Fourier) with domain alignment. The aligned features are encoded by a shared CLIP encoder with domain-specific expert modules. The resulting features are fused for classification and cosine alignment with text prompts. During training, a dual continual learning mechanism (RF-EWC and D-OGC) regulates parameter updates to prevent forgetting.}
\label{fig:framework}
\end{figure*}

Figure \ref{fig:framework} illustrates the overall architecture of Face-D\(^2\)CL. Given an input face image \(x_i\), the framework first transforms it into a \textbf{multi-domain synergistic representation} consisting of three complementary representations: Spatial feature \(x_i^S\), Wavelet feature \(x_i^W\), and Fourier feature \(x_i^F\). A batch-wise normalization aligns the distributions of \(x_i^W\) and \(x_i^F\) to that of \(x_i^S\), mitigating scale inconsistencies across domains. The aligned features are processed by a shared CLIP visual encoder with three domain-specific expert modules (LoRA), each dedicated to one domain. Features are fused by concatenation for classification and average pooling for text–visual alignment. Classification fusion feeds a lightweight binary classifier, while alignment fusion is used in a cosine alignment loss with fixed text prompts. During continual training, updates are regulated by a \textbf{dual continual learning mechanism} (RF-EWC and D-OGC) without storing historical data.

\subsection{Multi-Domain Synergistic Representation}

Existing deepfake detectors mainly use the Spatial domain for pixel-level inconsistencies and texture artifacts. They work well on known forgeries, yet single-domain features often limit generalization to unseen methods. Wavelet decomposition exposes multi-scale reconstruction artifacts from generative up-sampling that spatial cues alone miss; Fourier analysis reveals abnormal phase and magnitude patterns from frequency-domain synthesis, adding cues hard to see in pixel space. To overcome spatial-only limits, we extract and fuse Spatial, Wavelet, and Fourier information as \(\mathcal{F}(x) \rightarrow (f_{\text{cls}}, f_{\text{align}})\), where \(\mathcal{F}\) is the multi-domain module, \(f_{\text{cls}}\) the concatenated classification feature, and \(f_{\text{align}}\) the averaged text–visual alignment feature. The module has four parts below.

\paragraph{\textbf{\upshape Spatial, Wavelet, and Fourier Feature Extraction.}}
Given an input image \(x_i\), the Spatial representation preserves pixel-level forgery artifacts as \(x_i^S = x_i\). Wavelet and Fourier features use the modules from DFFreq \cite{yan2026dual}: a discrete wavelet transform yields low- and high-frequency coefficients, and high-frequency reconstruction keeps multi-scale details; an FFT separates magnitude and phase, a lightweight attention modulates phase, and the inverse transform gives a phase-enhanced reconstruction. Both modules join the multi-domain framework with the spatial branch under dual continual learning regularization.

\paragraph{\textbf{\upshape Domain Feature Alignment.}}
Multi-domain fusion is fragile when branch statistics differ: Wavelet / Fourier responses can overwhelm or be drowned by Spatial cues, so joint encoding becomes unbalanced. To restore balance, we re-center Wavelet and Fourier features onto the Spatial batch statistics. For a feature \(z\) (either \(x_i^W\) or \(x_i^F\)), the alignment is defined as:

\begin{equation}
\mathcal{N}(z; x_i^S) =
\frac{z - \mu_B(z)}{\sigma_B(z)} \cdot \sigma_B(x_i^S) + \mu_B(x_i^S),
\end{equation}

where \(\mu(\cdot)\) and \(\sigma(\cdot)\) are the mean and standard deviation computed over the batch. This inexpensive matching yields commensurate distributions, stabilizing parallel learning without discarding domain-specific content.

\paragraph{\textbf{\upshape Domain Expert Encoding.}}
A single frozen encoder cannot absorb all three domains without interference, yet allocating LoRA for each new forgery method or dataset is prohibitive as they grow. We therefore keep one CLIP visual encoder and attach a domain-wise expert module via Low-Rank Adaptation (LoRA) on query / value projections: for \(d\in\{S,W,F\}\), \(f_d = \mathcal{E}_d\bigl(x_i^{d'}\bigr)\) with \(\mathcal{E}_d\) the encoder under that expert. Shared pretrained weights provide a common semantic prior; separate low-rank experts specialize to each domain and can be updated across tasks without growing parameters per task, while later gradient projection curbs cross-task interference.

\paragraph{\textbf{\upshape Feature Fusion.}}
Complementary cues must be preserved for discrimination, yet a single fused vector is needed for language grounding. We thus use two pathways. Classification fusion concatenates the three features:

\begin{equation}
f_{\text{cls}} = \operatorname{concat}(f_S, f_W, f_F),
\end{equation}

which is then fed into a lightweight classifier for binary prediction, retaining domain-specific evidence. Alignment fusion averages the three features and L2-normalizes the result:

\begin{equation}
f_{\text{align}} =
\frac{1}{3}\sum_{d\in\{S,W,F\}} f_d,\qquad
\hat{f}_{\text{align}} = \frac{f_{\text{align}}}{\|f_{\text{align}}\|_2}.
\end{equation}

The latter serves as a neutral semantic anchor for a cosine alignment loss with fixed text prompts (e.g., ``real face'' and ``fake face''). Averaging, rather than learned weighting, avoids collapsing to one dominant branch and keeps the alignment objective stable across domains.

\subsection{Dual Continual Learning Mechanism}

Continual DeepFake detection requires balancing knowledge preservation with rapid adaptation. Conventional methods rely on data replay, which raises privacy and storage concerns, or on parameter regularization such as EWC, which applies uniform penalties without distinguishing real vs.\ fake importance. Gradient projection methods enforce orthogonality to prevent task interference, yet alone they miss a key challenge in face forgery detection: real faces remain compact and stable across tasks, whereas fake faces vary widely across forgery methods. Treating real and fake uniformly under such constraints therefore weakens real-face preservation and adaptation to new patterns; domain-expert updates may further disrupt the learned real-face manifold and miss emerging forgery cues, harming detection of both prior and new methods.

To overcome these, a replay-free dual continual learning mechanism combines RF-EWC to stabilize multi-domain features and D-OGC for low-rank plasticity.

\paragraph{\textbf{\upshape Real / Fake-aware Elastic Weight Consolidation (RF-EWC).}}
Here, RF denotes real / fake-aware Fisher estimation. Standard EWC computes a single Fisher information matrix for all parameters, treating importance uniformly across classes. In DeepFake detection, however, a parameter's importance for distinguishing real faces often differs from that for detecting specific forgery types. Two separate Fisher matrices are therefore estimated on each completed task \(k\):

\begin{equation}
\begin{aligned}
F^{\text{real}}_k &= \mathbb{E}_{x \sim \text{real}}\left[ \left( \nabla_{\theta} \log p(y=\text{real}|x,\theta) \right)^2 \right],\\
F^{\text{fake}}_k &= \mathbb{E}_{x \sim \text{fake}}\left[ \left( \nabla_{\theta} \log p(y=\text{fake}|x,\theta) \right)^2 \right].
\end{aligned}
\end{equation}

These Fisher matrices are estimated on the training data of the completed task, using the model weights with the best in-task validation performance. Regularization is applied to the frequency-domain feature extraction modules and the classification head to keep feature extraction stable across tasks. When learning a new task \(k+1\), a weighted penalty is applied:

\begin{equation}
\mathcal{L}_{\text{EWC}} = \sum_i \tfrac12\!\left( F^{\text{real}}_k(i) + F^{\text{fake}}_k(i) \right) (\theta_i - \theta_i^*)^2,
\end{equation}

where \(\theta_i\) are the current model parameters, \(\theta_i^*\) are the parameters after task \(k\), and \(F^{\text{real}}_k(i)\), \(F^{\text{fake}}_k(i)\) are the real / fake Fisher values at index \(i\). Unlike standard EWC, which builds one Fisher over mixed real / fake samples, RF-EWC separately computes real / fake Fisher weights to weight parameters by their combined importance.

\paragraph{\textbf{\upshape Domain-wise Orthogonal Gradient Constraint (D-OGC).}}
Here, D denotes domain-wise orthogonal projection. While RF-EWC stabilizes global parameters, domain experts capture task-specific forgery patterns. Unlike a single global orthogonal constraint over all adapters, D-OGC applies the same orthogonal projection separately to each Spatial / Wavelet / Fourier LoRA expert: for each expert, a gradient cache \(g_{\text{hist}}\) aggregates past-task directions, and the current gradient \(g\) is projected as
\begin{equation}
\label{eq:dogc}
g_{\parallel} = \frac{g^{\top} g_{\text{hist}}}{\|g_{\text{hist}}\|^2} g_{\text{hist}}, \qquad
g_{\perp} = g - g_{\parallel}, \qquad
\tilde{g} = g_{\perp}.
\end{equation}
\(\tilde{g}\) then updates that domain expert while preserving cached historical knowledge, without extra parameters or replay.

\paragraph{\textbf{\upshape Integrated Optimization and Overall Objective.}}
The synergy between RF-EWC and D-OGC allows the dual continual learning mechanism to achieve strong anti-forgetting while maintaining high plasticity. RF-EWC protects critical knowledge at the global parameter level, and D-OGC guides expert-module updates to remain orthogonal to historical knowledge, eliminating replay buffers. With a fixed number of domain experts, model size remains constant throughout continual learning. The final objective combines binary classification, RF-EWC, and a semantic alignment loss:

\begin{equation}
\mathcal{L}_{\text{loss}} = \mathcal{L}_{\text{BCE}} + \lambda_{\text{EWC}} \mathcal{L}_{\text{EWC}} + \lambda_{\text{align}} \mathcal{L}_{\text{align}},
\end{equation}
where \(\mathcal{L}_{\text{BCE}}\) is the binary cross-entropy loss on the classification head, \(\mathcal{L}_{\text{EWC}}\) is the RF-EWC regularization loss, and \(\mathcal{L}_{\text{align}}\) is the cosine alignment loss. The hyperparameters \(\lambda_{\text{EWC}}\) and \(\lambda_{\text{align}}\) balance these terms; we set \(\lambda_{\text{EWC}}=220\) with a linear schedule.

\begin{equation}
\mathcal{L}_{\text{align}} = 1 - \frac{1}{B} \sum_{i=1}^B \hat{f}_{\text{align}}^{(i)} \cdot \mathbf{t}_{y_i},
\end{equation}
with \(\mathbf{t}_{y_i}\) being the normalized text feature of the prompt corresponding to the ground-truth label.

The combination of Multi-Domain Synergistic Representation across Spatial, Wavelet, and Fourier domains and Dual Continual Learning constraints yields a practical framework with strong generalization to unseen forgery types and robust adaptation to new tasks.

\section{Experiments}

We evaluate continual learning, generalization to unseen forgery domains, and robustness under common degradations; ablation studies, implementation details, and further continual learning analyses appear in the Appendix.

\begin{table*}[t]
\caption{Performance comparison on dataset-incremental (left) and forgery-type incremental (right) protocols. Each method is shown across four stages, corresponding to the performance on seen datasets after each task. Results are reported in terms of accuracy (ACC) for Protocol 1 and area under the ROC curve (AUC) for Protocol 2. Best results are shown in bold.}
\label{tab:main_results}
\centering
\scriptsize
\setlength{\tabcolsep}{2.4pt}
\renewcommand{\arraystretch}{0.80}
\resizebox{\textwidth}{!}{%
\begin{tabular}{c c c c c c c c c c c c c c}
\toprule
\multirow{2}{*}{Method} &
\multirow{2}{*}{Venue} &
\multirow{2}{*}{No-Replay} &
\multirow{2}{*}{Stage} &
\multicolumn{5}{c}{Protocol 1 (Dataset Incremental)} &
\multicolumn{5}{c}{Protocol 2 (Forgery Type Incremental)} \\
\cmidrule(lr){5-9} \cmidrule(lr){10-14}
& & & & FF++ & DFDCP & DFD & CDF2 & Avg & Hybrid & FR & FS & EFS & Avg \\
\midrule

\multicolumn{14}{c}{\textit{Offline (non-incremental) methods}} \\
\multirow{4}{*}{DFD-FCG} &
\multirow{4}{*}{CVPR'25} &
\multirow{4}{*}{/} &
Task1 & 0.997 & -      & -      & -      & 0.997         & 0.994      & -      & -      & -      & 0.994        \\
& & & Task2 & 0.995 & 0.957 & -      & -      & 0.976         & 0.985      & 0.952      & -      & -      & 0.967         \\
& & & Task3 & 0.995 & 0.942 & 0.963 & -      & 0.967         & 0.976      & 0.959      & 0.743      & -      & 0.893         \\
& & & Task4 & 0.995 & 0.941 & 0.960 & 0.934 & 0.956 & 0.975      & 0.977      & 0.673      & 0.999      & 0.906 \\

\multirow{4}{*}{DFFreq} &
\multirow{4}{*}{TIFS'26} &
\multirow{4}{*}{/} &
Task1 & 0.997 & -      & -      & -      & 0.997 & 0.570      & -      & -      & -      & 0.570      \\
& & & Task2 & 0.387 & 0.894 & -      & -      & 0.640 & 0.535      & 0.999      & -      & -      & 0.767      \\
& & & Task3 & 0.903 & 0.442 & 0.994 & -      & 0.780 & 0.578      & 0.999      & 0.885      & -      & 0.821      \\
& & & Task4 & 0.836 & 0.568 & 0.781 & 0.839 & 0.756 & 0.505      & 0.999      & 0.872      & 0.999      & 0.844      \\

\multicolumn{14}{c}{\textit{Continual learning methods}} \\
\multirow{4}{*}{DFIL} &
\multirow{4}{*}{ACM MM'23} &
\multirow{4}{*}{$\times$} &
Task1 & 0.977 & -      & -      & -      & 0.977 & 0.949 & -      & -      & -      & 0.949 \\
& & & Task2 & 0.974 & 0.829 & -      & -      & 0.902 & 0.871 & 0.997 & -      & -      & 0.934 \\
& & & Task3 & 0.942 & 0.777 & 0.807 & -      & 0.842 & 0.827 & 0.967 & 0.989 & -      & 0.928 \\
& & & Task4 & 0.874 & 0.681 & 0.742 & 0.884 & 0.795 & 0.821 & 0.959 & 0.973 & 0.999 & 0.938 \\

\multirow{4}{*}{SUR-LID} &
\multirow{4}{*}{CVPR'25} &
\multirow{4}{*}{$\times$} &
Task1 & 0.994 & -      & -      & -      & 0.994 & 0.852 & -      & -      & -      & 0.852 \\
& & & Task2 & 0.994 & 0.926 & -      & -      & 0.960 & 0.861 & 0.973 & -      & -      & 0.917 \\
& & & Task3 & 0.987 & 0.932 & 0.981 & -      & 0.967 & 0.842 & 0.964 & 0.620 & -      & 0.809 \\
& & & Task4 & 0.987 & 0.936 & 0.974 & 0.942 & 0.959 & 0.861 & 0.968 & 0.660 & 0.999 & 0.872 \\

\multirow{4}{*}{SAIDO} &
\multirow{4}{*}{CVPR'26} &
\multirow{4}{*}{$\checkmark$} &
Task1 & 0.990 & -      & -      & -      & 0.990 & 0.999      & -      & -      & -      & \textbf{0.999}      \\
& & & Task2 & 0.981 & 0.839 & -      & -      & 0.910 & 0.990      & 0.925      & -      & -      & 0.958      \\
& & & Task3 & 0.990 & 0.852 & 0.974 & -      & 0.939 & 0.968      & 0.967      & 0.743      & -      & 0.893      \\
& & & Task4 & 0.971 & 0.858 & 0.977 & 0.858 & 0.916 & 0.987      & 0.914      & 0.340      & 0.975      & 0.804      \\

\rowcolor{cyan!10}
\cellcolor{cyan!10} &
\cellcolor{cyan!10} &
\cellcolor{cyan!10} &
Task1 & 0.998 & -      & -      & -      & \textbf{0.998} & 0.998  & -      & -      & -      & 0.998  \\
\rowcolor{cyan!10}
\cellcolor{cyan!10} &
\cellcolor{cyan!10} &
\cellcolor{cyan!10} &
Task2 & 0.996 & 0.962 & -      & -      & \textbf{0.979} & 0.992  & 0.997  & -      & -      & \textbf{0.995}  \\
\rowcolor{cyan!10}
\cellcolor{cyan!10} &
\cellcolor{cyan!10} &
\cellcolor{cyan!10} &
Task3 & 0.998 & 0.944 & 0.980 & -      & \textbf{0.974} & 0.990 & 0.989 & 0.989 & -      & \textbf{0.989}  \\
\rowcolor{cyan!10}
\multirow{-4}{*}{\cellcolor{cyan!10}\textbf{Face-D\(^2\)CL}} &
\multirow{-4}{*}{\cellcolor{cyan!10}-} &
\multirow{-4}{*}{\cellcolor{cyan!10}$\checkmark$} &
Task4 & 0.998 & 0.952 & 0.964 & 0.952 & \textbf{0.967} & 0.978 &0.980 & 0.988 & 0.999 & \textbf{0.986} \\
\bottomrule
\end{tabular}
}
\end{table*}

\begin{figure*}[t]
    \centering
    \includegraphics[width=1.00\linewidth,height=0.30\textheight,keepaspectratio]{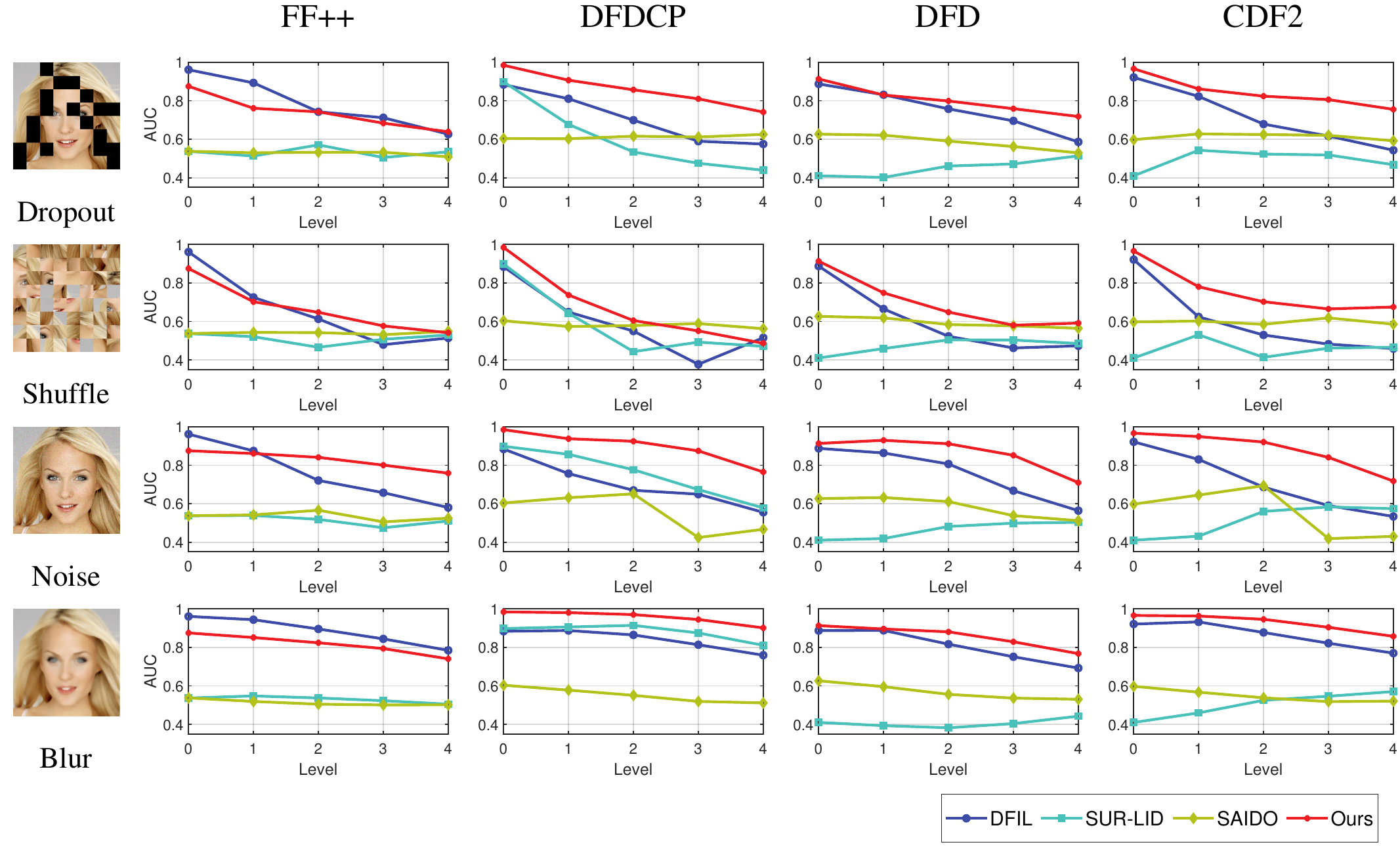}
    \caption{Robustness comparison of different methods under unseen perturbations based on Protocol 1. Average AUC across all test sets is reported under four perturbation types, namely block-wise dropout, grid shuffle, Gaussian noise, and median blur, each evaluated at five intensity levels.}
    \label{fig:robustness}
\end{figure*}

\subsection{Experimental Settings}

\paragraph{\textbf{\upshape Datasets.}}
The evaluation employs a diverse collection of face deepfake datasets covering three fundamental forgery categories: Face-Swapping (FS), Face-Reenactment (FR), and Entire Face Synthesis (EFS) \cite{yan2024df40}. Specifically, three classical FS datasets are used for Protocol 1: Celeb-DF v2 (CDF2) \cite{li2020face}, Deepfake Detection Challenge Preview (DFDCP) \cite{dolhansky2019deepfake}, and Deepfake Detection (DFD) \cite{dufour2019contributing}. FaceForensics++ (FF++) \cite{rossler2019faceforensics++} contains four forgery methods covering both FS and FR, and is therefore treated as a dataset with Hybrid forgery categories. For Protocol 2, datasets with more diverse forgery categories and techniques released in recent years are further employed, including MCNet \cite{hong2023implicit} (FR), BlendFace \cite{shiohara2023blendface} (FS), and StyleGAN3 \cite{karras2021alias} (EFS) from DF40 \cite{yan2024df40}.

\paragraph{\textbf{\upshape Protocols.}}
Two incremental protocols are adopted following prior works \cite{pan2023dfil, tian2024dynamic}. 
\textbf{Protocol 1 (Dataset Incremental)}: the task sequence is FF++ → DFDCP → DFD → CDF2. The first task uses the full training set, while for each subsequent task 25 fake videos are randomly sampled to simulate the scenario where only limited new data is available.
\textbf{Protocol 2 (Forgery Type Incremental)}: the task sequence is Hybrid (FF++ containing multiple forgery types) → FR (MCNet, face-reenactment) → FS (BlendFace, face-swapping) → EFS (StyleGAN3, entire face synthesis). Here the real data remain identical across tasks, while only the fake domain changes. All datasets are strictly split at the video level to prevent cross-task contamination. Frames are cropped to face regions and normalized.

\paragraph{\textbf{\upshape Evaluation Metrics.}}
Standard metrics in continual learning and deepfake detection are used:
\begin{itemize}
    \item \textbf{Accuracy (ACC)}: the ratio of correctly predicted samples, with thresholds selected per dataset on validation set.
    \item \textbf{Average Accuracy (AA)}: \(\text{AA} = \frac{1}{T}\sum_{i=1}^{T} a_{T,i}\), where \(a_{T,i}\) is the accuracy on task \(i\) after all tasks.
    \item \textbf{Average Forgetting (AF)}: \(\text{AF} = \frac{1}{T-1}\sum_{i=1}^{T-1} (a_{i,i} - a_{T,i})\), where \(a_{i,i}\) is the accuracy after training task \(i\).
    \item \textbf{Area Under the ROC Curve (AUC)}: a threshold-independent metric used for evaluating generalization to unseen forgery datasets and robustness under perturbations.
\end{itemize}

\paragraph{\textbf{\upshape Compared Methods.}}
The proposed method is compared with both offline (non-incremental) and continual learning methods. All methods are re-implemented under the same settings for fair comparison.
\begin{itemize}
    \item \textbf{Offline (non-incremental)}: DFD-FCG \cite{han2025towards} and DFFreq \cite{yan2026dual}, incrementally fine-tuned on each new dataset at every stage without continual-learning mechanisms.
    \item \textbf{Continual learning}: replay-based methods including DFIL \cite{pan2023dfil} and SUR-LID \cite{cheng2025stacking}, as well as the replay-free method SAIDO \cite{hu2026saido}.
\end{itemize}

\subsection{Performance Evaluation on Continual Learning Tasks}

We evaluate continual learning performance on both protocols against the compared methods under the same settings. Table~\ref{tab:main_results} reports the accuracy on each dataset after each task, along with the overall average accuracy (AA). The results demonstrate that the proposed method achieves the highest AA across both protocols, substantially outperforming existing approaches. Notably, existing methods struggle to perform well in Protocol 2, where forgery methods are diverse and real images remain in the same domain across tasks. In this scenario, detectors are more prone to overriding previously learned forgery information because forgery-irrelevant features are consistent across different forgery types, making the feature representations more similar. This implies that existing methods may not fully capture the specific forgery patterns of each task, leading to catastrophic forgetting. In contrast, the proposed Face-D\(^2\)CL effectively mitigates this issue through its dual continual learning mechanism, which preserves real-face representations while flexibly adapting to new forgery patterns without relying on data replay.

\subsection{Generalization Ability}

To evaluate generalization to unseen forgery domains, the model trained on Protocol 1 is tested on three additional datasets: DF40 \cite{yan2024df40}, UADFV \cite{yang2019exposing}, and WildDeepfake \cite{zi2020wilddeepfake}. Each test dataset contains approximately 2,000 samples, with balanced real and fake images. Table~\ref{tab:generalization} reports the frame-level AUC on these datasets.

\begin{table}[t]
\caption{Generalization under Protocol~1 (AUC).}
\label{tab:generalization}
\centering
\small
\setlength{\tabcolsep}{4pt}
\renewcommand{\arraystretch}{0.85}
\begin{tabular}{lcccc}
\toprule
Method & DF40 & UADFV & WildDeepfake & AA \\
\midrule
DFD-FCG      & 0.688 & 0.810 & 0.673 & 0.724 \\
DFFreq         & 0.668 & 0.645 & 0.681 & 0.665 \\
DFIL           & 0.680 & 0.961 & 0.708 & 0.783 \\
SUR-LID  & 0.773 & 0.873 & 0.692 & 0.779 \\
SAIDO          & 0.780 & 0.905 & 0.705 & 0.797 \\
\rowcolor{cyan!10}
\textbf{Face-D\(^2\)CL}           & \textbf{0.846} & \textbf{0.971} & \textbf{0.812} & \textbf{0.876} \\
\bottomrule
\end{tabular}
\end{table}

The proposed method consistently outperforms prior approaches across all three unseen datasets. Compared to SAIDO, Face-D\(^2\)CL achieves absolute improvements of 6.6\%, 6.6\%, and 10.7\% on DF40, UADFV, and WildDeepfake, respectively, leading to an overall average AUC improvement of 7.9\%. This demonstrates the effectiveness of multi-domain synergistic representation. The improvement stems from two factors. First, the joint extraction of Spatial, Wavelet, and Fourier features captures complementary forgery artifacts that are more domain-agnostic than single-domain features, enabling transferable detection on novel forgery types. Second, the dual continual learning mechanism preserves the learned representations during incremental training, mitigating catastrophic forgetting that would otherwise compromise generalization.

\subsection{Robustness Evaluation}

To evaluate model resilience against unseen image degradations, the robustness of different methods is assessed under four types of perturbations. Specifically, the model trained on Protocol 1 after completing all four tasks is used as the baseline for robustness testing. Four types of perturbations are considered: block-wise dropout (mask ratios of 0.1, 0.2, 0.3, and 0.4), grid shuffle (patch sizes of 2, 4, 8, and 12), Gaussian noise (standard deviations of 0.01, 0.02, 0.04, and 0.08), and median blur (kernel sizes of 3, 5, 7, and 9). Each perturbation is applied at four intensity levels, with the unperturbed (Level 0) case serving as the reference, resulting in five evaluation points per perturbation. Figure~\ref{fig:robustness} illustrates the average AUC across all test sets, presented as line charts for clear comparison across methods and perturbation levels.

The proposed method generally achieves competitive or superior AUC across most datasets and perturbation types, while exhibiting relatively smaller performance degradation as perturbation intensity increases. For block-wise dropout and grid shuffle, which introduce severe structural distortions, the proposed method maintains a higher AUC than the competing approaches across the majority of intensity levels, with a notably slower decline. Under Gaussian noise, which simulates typical sensor noise, the proposed method sustains relatively strong detection performance even at the highest noise level, whereas other methods experience sharper drops in AUC on datasets such as DFDCP and CDF2. Under median blur, a relatively mild perturbation, the proposed method still yields favorable AUC across most datasets and intensities.

Overall, the observed robustness can be attributed to the multi-domain synergistic representation and the dual continual learning mechanism. By jointly extracting complementary forgery traces from spatial, wavelet, and Fourier domains, the model learns a more stable and discriminative latent space that is less sensitive to image degradations. Meanwhile, D-OGC and RF-EWC preserve the learned representations during incremental training, preventing catastrophic forgetting that would otherwise compromise robustness.

\section{Conclusion}

This paper presents Face-D\(^2\)CL, a novel framework for continual DeepFake detection that integrates multi-domain synergistic representation with a dual continual learning mechanism. The multi-domain synergistic representation extracts complementary forgery traces from spatial, wavelet, and Fourier domains, providing a comprehensive feature space that enhances generalization across diverse face forgery methods. The dual continual learning mechanism, comprising RF-EWC and D-OGC, operates without data replay: RF-EWC preserves global parameter stability through real / fake Fisher information, while D-OGC ensures domain-expert updates remain orthogonal to historical gradient directions, enabling flexible adaptation. Their synergy achieves a dynamic balance between stability and plasticity. Extensive experiments on both dataset-incremental and forgery-type incremental protocols demonstrate state-of-the-art performance, with significant reductions in average forgetting and improved generalization to unseen forgery domains.

\clearpage
\bibliography{references}
\clearpage
\appendix

\section{Experimental Setup}

\subsection{Implementation Details}
The CLIP ViT-L/14 model is adopted as the backbone, pre-trained on large-scale vision-language data. Three independent domain expert modules (LoRA, rank \(r=4\), \(\alpha=16\)) are injected into the query and value projections, each dedicated to one domain (Spatial, Wavelet, Fourier). The fused feature dimension is \(3\times768=2304\), followed by a two-layer classifier with GELU and dropout (0.5). Training uses the Adam optimizer (learning rate \(1.2\times10^{-5}\), batch size 48) for 25 epochs per task. The RF-EWC coefficient is \(\lambda_{\text{EWC}}=220\) with a linear schedule. The alignment loss weight is \(\lambda_{\text{align}}=0.5\). Unless noted otherwise, each appendix setting, including its full-model reference, is trained independently under this recipe.

\subsection{Fair Comparison with Replay-based Methods}
Under Protocols~1 and~2, all non-method-specific training settings are kept identical; baselines retain their original backbones and method-specific configurations, including the buffer settings of DFIL and SUR-LID. Face-D\(^2\)CL requires no replay buffer, reducing memory overhead and privacy risk and making it more practical for real-world deployment.

\section{Ablation Studies}

This appendix isolates the contribution of each major design choice in Face-D\(^2\)CL. We first examine RF-EWC and the frequency branches, followed by the text–visual alignment loss, then individual Spatial / Wavelet / Fourier experts and the dual continual learning components (D-OGC, RF-EWC, and standard Fisher). Cross-branch CKA and D-OGC numerical diagnostics are reported in Appendix~\ref{sec:mechanism}.

\subsection{Progressive Component Ablation}
We ablate RF-EWC under Protocols~1 and~2 (w/o RF-EWC). As shown in Table~\ref{tab:ablation_core}, in Protocol 1 this leads to a 5.5\% decrease in average accuracy and a 12.7\% increase in forgetting, confirming the effectiveness of RF-EWC in maintaining stability under dataset-incremental settings. In Protocol 2, the variant results in a 0.4\% decrease in average AUC and a 1.1\% increase in forgetting. This indicates that RF-EWC improves performance and reduces forgetting across both protocols, with a particularly pronounced effect on stability when the data distribution shifts significantly.

\begin{center}
\captionof{table}{Progressive ablation of RF-EWC and frequency branches (Protocol~1: ACC; Protocol~2: AUC).}
\label{tab:ablation_core}
\small
\setlength{\tabcolsep}{3pt}
\renewcommand{\arraystretch}{0.90}
\begin{tabular}{lcccc}
\toprule
\multirow{2}{*}{Configuration} & \multicolumn{2}{c}{Protocol 1} & \multicolumn{2}{c}{Protocol 2} \\
\cmidrule(lr){2-3} \cmidrule(lr){4-5}
& AA & AF & AA & AF \\
\midrule
w/o RF-EWC & 0.925 & 0.133 & 0.988 & 0.024 \\
w/o RF-EWC \& Freq & 0.893 & 0.282 & 0.982 & 0.034 \\
\rowcolor{cyan!10}
\textbf{Face-D\(^2\)CL} & \textbf{0.980} & \textbf{0.006} & \textbf{0.992} & \textbf{0.013} \\
\bottomrule
\end{tabular}
\end{center}

We further jointly remove RF-EWC and the Wavelet / Fourier branches (w/o RF-EWC \& Freq). Compared to the w/o RF-EWC variant, this leads to a 3.2\% drop in average accuracy on Protocol 1 and a 0.6\% drop on Protocol 2, while the forgetting rate increases by 0.149 on Protocol 1. These results demonstrate that the multi-domain synergistic representation contributes substantially to discriminability and robustness, as its removal consistently degrades accuracy across both protocols. The progressive degradation observed across the two ablation variants underscores the complementary nature of the two components. RF-EWC anchors the model to prior knowledge, ensuring stability, while the multi-domain synergistic representation enriches the feature space with complementary forgery cues, enhancing generalization. Their synergy enables the full model to achieve balanced and consistent performance across both protocols, as reflected by strong overall accuracy and low forgetting.

\subsection{Alignment Loss Ablation}
The text–visual alignment loss \(\mathcal{L}_{\text{align}}\) encourages domain-invariant feature learning by aligning fused features with their corresponding fixed text prompts. As shown in Table~\ref{tab:ablation_align}, removing this loss on Protocol~1 leads to a clear drop in average accuracy, confirming its effectiveness in maintaining semantic consistency across domains.

\begin{center}
\captionof{table}{Ablation on alignment loss under Protocol~1 (ACC).}
\label{tab:ablation_align}
\small
\setlength{\tabcolsep}{3pt}
\renewcommand{\arraystretch}{0.90}
\begin{tabular}{lccccc}
\toprule
Setting & FF++ & DFDCP & DFD & CDF2 & AA \\
\midrule
Without alignment & 0.987 & 0.953 & 0.970 & 0.944 & 0.964 \\
\rowcolor{cyan!10}
\textbf{With alignment} & \textbf{0.998} & \textbf{0.953} & \textbf{0.972} & \textbf{0.952} & \textbf{0.969} \\
\bottomrule
\end{tabular}
\end{center}

\subsection{Multi-domain Branch Ablation}
To verify that each of the Spatial, Wavelet, and Fourier branches contributes non-redundant cues, we remove one branch at a time on Protocol~1. As shown in Table~\ref{tab:ablation_multidomain}, disabling any branch lowers AA by at least 4.8\%. Cross-branch representational complementarity is further examined in Section~\ref{sec:cka}.

\begin{center}
\captionof{table}{Multi-domain branch ablation on Protocol~1 (ACC).}
\label{tab:ablation_multidomain}
\small
\setlength{\tabcolsep}{4pt}
\renewcommand{\arraystretch}{0.90}
\begin{tabular}{@{}lcc@{}}
\toprule
Config & AA & AF \\
\midrule
w/o Spatial & 0.940 & 0.010 \\
w/o Wavelet & 0.899 & 0.011 \\
w/o Fourier & 0.903 & 0.013 \\
\rowcolor{cyan!10}
\textbf{Face-D\(^2\)CL} & \textbf{0.988} & \textbf{0.007} \\
\bottomrule
\end{tabular}
\end{center}

\subsection{Dual Continual Learning Ablation}
To isolate the dual continual learning modules, we remove D-OGC, remove RF-EWC, or replace real / fake Fisher with standard Fisher on Protocol~1. As shown in Table~\ref{tab:ablation_dualcl}, without D-OGC AA drops 2.9\%; without RF-EWC AF rises sharply; standard Fisher costs 8.0\% AA and $2.1\times$ AF versus the full model.

\begin{center}
\captionof{table}{Dual continual learning ablation on Protocol~1 (ACC).}
\label{tab:ablation_dualcl}
\small
\setlength{\tabcolsep}{4pt}
\renewcommand{\arraystretch}{0.90}
\begin{tabular}{@{}lcc@{}}
\toprule
Config & AA & AF \\
\midrule
w/o D-OGC & 0.959 & 0.005 \\
w/o RF-EWC & 0.925 & 0.133 \\
Std Fisher & 0.908 & 0.015 \\
\rowcolor{cyan!10}
\textbf{Face-D\(^2\)CL} & \textbf{0.988} & 0.007 \\
\bottomrule
\end{tabular}
\end{center}

\section{Additional Results}

\subsection{Task-Order Robustness}
Continual detectors may be sensitive to the order of incoming datasets. The default Protocol~1 order is FF++ $\rightarrow$ DFDCP $\rightarrow$ DFD $\rightarrow$ CDF2; we also evaluate Order~2 that swaps CDF2 and DFDCP (FF++ $\rightarrow$ CDF2 $\rightarrow$ DFD $\rightarrow$ DFDCP). As shown in Table~\ref{tab:task_order_compare}, Face-D\(^2\)CL's final-task AA differs by only 0.005 from the default order, while baselines fluctuate more. This suggests that RF-EWC preserves globally important real / fake parameters while D-OGC keeps domain-expert updates from overwriting earlier forgery-specific directions when the task sequence changes.

\subsection{Detailed Robustness under Perturbations}
Figure~\ref{fig:robustness} summarizes average AUC trends under perturbations. Tables~\ref{tab:robustness_dropout_shuffle} and~\ref{tab:robustness_noise_blur} further break down Protocol~1 results by dataset under block-wise dropout, grid shuffle, Gaussian noise, and median blur at five intensity levels. Face-D\(^2\)CL maintains favorable AUC with slower degradation as intensity increases. At Level~4 it retains average AUC of 0.714 under block-wise dropout and 0.573 under grid shuffle, remaining higher than competing methods; under Gaussian noise and median blur it likewise degrades more gracefully. The per-dataset breakdowns confirm robustness, which may be beneficial under compression, occlusion-like dropout, and mild geometric scrambling in social-media redistribution.

\subsection{Computational Efficiency}
Because Face-D\(^2\)CL is replay-free, we report computational cost under fair Protocol settings and the training overhead of RF-EWC / D-OGC. Face-D\(^2\)CL takes 73.49\,s for training and 9.77\,s for testing, comparable to SAIDO and well below DFD-FCG (214.10\,s for training and 20.16\,s for testing); it requires 467.76 / 233.88\,GFLOPs for training / testing, higher than SAIDO and SUR-LID owing to the multi-domain branches, but Face-D\(^2\)CL uses 24.18\,GB peak training memory, which remains in a similar range to SAIDO, and training on $2\times$RTX~4090 is acceptable given the accuracy gains. As shown in Table~\ref{tab:ewc_ogc_overhead}, enabling RF-EWC adds only 0.05\,s per training epoch with no peak-memory increase, and enabling D-OGC adds 2.83\,s per epoch and 0.05\,GB; both overheads are training-only and do not change inference cost.

\subsection{Real-face Manifold Preservation}
In forgery-type incremental learning, fake faces change sharply across generators while real faces should remain a compact, stable cluster. RF-EWC is designed to protect real- and fake-critical parameters separately; if effective, continual adaptation should less severely fragment the real-face manifold. On Protocol~2 we embed test real faces and measure $k$-NN subset pseudo-separation: for each real sample we take its $k{=}15$ nearest neighbors and report the fraction from the same subset (lower is better). As shown in Table~\ref{tab:manifold}, Face-D\(^2\)CL obtains the lowest score of 0.279, outperforming SUR-LID (0.309) and SAIDO (0.366) among continual-learning methods and also improving over DFFreq (0.288). This indicates that Face-D\(^2\)CL better preserves real-face structure under continual learning, consistent with RF-EWC's real / fake-aware Fisher estimation.

\subsection{Longer Task Sequences and Recent Forgeries}
To evaluate the effectiveness of Face-D\(^2\)CL over longer task streams, we append SimSwap, FOMM, SiT-XL/2, and CollabDiff after FF++, DFDCP, DFD, and CDF2 to form an eight-task sequence. As shown in Table~\ref{tab:eight_task}, after Task~8 Face-D\(^2\)CL reaches 94.6\% / 2.5\% AA / AF, outperforming the SOTA replay-free baseline SAIDO by 0.8\% AA and 0.1\% lower AF, and the SOTA replay-based baseline SUR-LID by 4.4\% AA and 5.0\% lower AF. The sustained AA / AF indicates that the dual continual learning mechanism scales beyond the short four-task setting without requiring additional replay memory.

\subsection{Generalization to General AIGC Detection}
To test transfer beyond face-centric streams, we evaluate a nine-task general AIGC continual benchmark spanning diverse generators (e.g., ADM, GLIDE, ProGAN, SAGAN, BigGAN, wukong, SD1.5, Midjourney, VQDM). As shown in Table~\ref{tab:aigc9}, Face-D\(^2\)CL achieves 0.959 / 0.024 AA / AF, outperforming SAIDO by 3.0\% AA and 4.7\% lower AF. The larger AF gap suggests that RF-EWC and D-OGC remain beneficial when the incremental stream leaves the face regime and spans heterogeneous generative priors.

\subsection{Run-to-Run Stability}
To assess run-to-run stability, we repeat the full model five times under the same experimental configuration with different random seeds; the runs achieve a mean final performance of \(0.978\) with a standard deviation of \(0.010\), indicating good numerical stability across runs. Within each table, the full model and its ablated variants share the same fixed random seed, whereas independently retrained tables use different seeds.

\clearpage
\twocolumn[{%
\begin{minipage}{\textwidth}
\centering
\captionof{table}{Order~2 dataset-incremental comparison (ACC). Sequence: FF++ $\rightarrow$ CDF2 $\rightarrow$ DFD $\rightarrow$ DFDCP.}
\label{tab:task_order_compare}
\scriptsize
\setlength{\tabcolsep}{1.2pt}
\renewcommand{\arraystretch}{1.00}
\resizebox{\textwidth}{!}{%
\begin{tabular}{c c c c c c c c c}
\toprule
\multirow{2}{*}{Method} &
\multirow{2}{*}{Venue} &
\multirow{2}{*}{No-Replay} &
\multirow{2}{*}{Stage} &
\multicolumn{5}{c}{alternative order} \\
\cmidrule(lr){5-9}
& & & & FF++ & CDF2 & DFD & DFDCP & Avg \\
\midrule
\multicolumn{9}{c}{\textit{Offline (non-incremental) methods}} \\
\multirow{4}{*}{DFD-FCG} &
\multirow{4}{*}{CVPR'25} &
\multirow{4}{*}{/} &
Task1 & 0.9956 & -      & -      & -      & 0.9956 \\
& & & Task2 & 0.9989 & 0.9581 & -      & -      & 0.9785 \\
& & & Task3 & 0.9978 & 0.9323 & 0.8677 & -      & 0.9326 \\
& & & Task4 & 0.9963 & 0.9032 & 0.8742 & 0.8613 & 0.9088 \\
\multirow{4}{*}{DFFreq} &
\multirow{4}{*}{TIFS'26} &
\multirow{4}{*}{/} &
Task1 & 0.9935 & -      & -      & -      & 0.9935 \\
& & & Task2 & 0.8484 & 0.8323 & -      & -      & 0.8404 \\
& & & Task3 & 0.8935 & 0.5129 & 0.9935 & -      & 0.8000 \\
& & & Task4 & 0.7355 & 0.7774 & 0.7645 & 0.8903 & 0.7919 \\
\multicolumn{9}{c}{\textit{Continual learning methods}} \\
\multirow{4}{*}{DFIL} &
\multirow{4}{*}{ACM MM'23} &
\multirow{4}{*}{$\times$} &
Task1 & 0.9065 & -      & -      & -      & 0.9065 \\
& & & Task2 & 0.8581 & 0.9290 & -      & -      & 0.8936 \\
& & & Task3 & 0.7355 & 0.8903 & 0.9258 & -      & 0.8505 \\
& & & Task4 & 0.7839 & 0.8452 & 0.7774 & 0.9645 & 0.8428 \\
\multirow{4}{*}{SUR-LID} &
\multirow{4}{*}{CVPR'25} &
\multirow{4}{*}{$\times$} &
Task1 & 0.9903 & -      & -      & -      & 0.9903 \\
& & & Task2 & 0.9903 & 0.9516 & -      & -      & 0.9710 \\
& & & Task3 & 0.9935 & 0.9323 & 0.9774 & -      & 0.9677 \\
& & & Task4 & 0.9903 & 0.9194 & 0.9774 & 0.9419 & 0.9573 \\
\multirow{4}{*}{SAIDO} &
\multirow{4}{*}{CVPR'26} &
\multirow{4}{*}{$\checkmark$} &
Task1 & 0.9935 & -      & -      & -      & 0.9935 \\
& & & Task2 & 0.9871 & 0.8548 & -      & -      & 0.9210 \\
& & & Task3 & 0.9839 & 0.8645 & 0.9774 & -      & 0.9419 \\
& & & Task4 & 0.9806 & 0.8613 & 0.9710 & 0.8355 & 0.9121 \\
\rowcolor{cyan!10}
\cellcolor{cyan!10} &
\cellcolor{cyan!10} &
\cellcolor{cyan!10} &
Task1 & 0.9980 & -      & -      & -      & \textbf{0.9980} \\
\rowcolor{cyan!10}
\cellcolor{cyan!10} &
\cellcolor{cyan!10} &
\cellcolor{cyan!10} &
Task2 & 0.9960 & 0.9720 & -      & -      & \textbf{0.9840} \\
\rowcolor{cyan!10}
\cellcolor{cyan!10} &
\cellcolor{cyan!10} &
\cellcolor{cyan!10} &
Task3 & 0.9960 & 0.9600 & 0.9620 & -      & \textbf{0.9727} \\
\rowcolor{cyan!10}
\multirow{-4}{*}{\cellcolor{cyan!10}\textbf{Face-D\(^2\)CL}} &
\multirow{-4}{*}{\cellcolor{cyan!10}-} &
\multirow{-4}{*}{\cellcolor{cyan!10}$\checkmark$} &
Task4 & 0.9900 & 0.9700 & 0.9580 & 0.9700 & \textbf{0.9720} \\
\bottomrule
\end{tabular}
}

\end{minipage}
}]

\twocolumn[{%
\begin{minipage}{\textwidth}
\centering
\captionof{table}{Robustness results under block-wise dropout and grid shuffle perturbations. Each perturbation is applied at five intensity levels (Level 0: unperturbed; Levels 1--4: increasing strength). Results are reported in AUC.}
\label{tab:robustness_dropout_shuffle}
\tiny
\setlength{\tabcolsep}{1.4pt}
\renewcommand{\arraystretch}{1.00}
\resizebox{\textwidth}{!}{%
\begin{tabular}{c c c c c c c c c c c c c c}
\toprule
\multirow{2}{*}{Method} &
\multirow{2}{*}{Venue} &
\multirow{2}{*}{No-Replay} &
\multirow{2}{*}{Level} &
\multicolumn{5}{c}{Block-wise Dropout} &
\multicolumn{5}{c}{Grid Shuffle} \\
\cmidrule(lr){5-9} \cmidrule(lr){10-14}
& & & & FF++ & DFDCP & DFD & CDF2 & Avg & FF++ & DFDCP & DFD & CDF2 & Avg \\
\midrule
\multirow{5}{*}{DFIL} &
\multirow{5}{*}{ACM MM'23} &
\multirow{5}{*}{$\times$} &
0 & \textbf{0.962} & 0.884 & 0.888 & 0.922 & 0.914 & \textbf{0.962} & 0.884 & 0.888 & 0.922 & 0.914 \\
& & & 1 & \textbf{0.893} & 0.811 & \textbf{0.831} & 0.823 & \textbf{0.840} & \textbf{0.725} & 0.649 & 0.665 & 0.624 & 0.666 \\
& & & 2 & \textbf{0.743} & 0.699 & 0.758 & 0.679 & 0.720 & 0.613 & 0.550 & 0.523 & 0.530 & 0.554 \\
& & & 3 & \textbf{0.712} & 0.590 & 0.696 & 0.617 & 0.654 & 0.479 & 0.377 & 0.462 & 0.482 & 0.450 \\
& & & 4 & 0.626 & 0.575 & 0.586 & 0.543 & 0.583 & 0.514 & 0.516 & 0.473 & 0.459 & 0.491 \\

\multirow{5}{*}{SUR-LID} &
\multirow{5}{*}{CVPR'25} &
\multirow{5}{*}{$\times$} &
0 & 0.537 & 0.899 & 0.410 & 0.409 & 0.564 & 0.537 & 0.899 & 0.410 & 0.409 & 0.564 \\
& & & 1 & 0.513 & 0.677 & 0.401 & 0.543 & 0.534 & 0.521 & 0.643 & 0.459 & 0.531 & 0.539 \\
& & & 2 & 0.571 & 0.534 & 0.461 & 0.523 & 0.522 & 0.466 & 0.443 & 0.504 & 0.413 & 0.457 \\
& & & 3 & 0.505 & 0.475 & 0.471 & 0.518 & 0.492 & 0.507 & 0.493 & 0.503 & 0.461 & 0.491 \\
& & & 4 & 0.535 & 0.439 & 0.514 & 0.467 & 0.489 & 0.530 & 0.471 & 0.485 & 0.466 & 0.488 \\

\multirow{5}{*}{SAIDO} &
\multirow{5}{*}{CVPR'26} &
\multirow{5}{*}{$\checkmark$} &
0 & 0.537 & 0.604 & 0.627 & 0.598 & 0.592 & 0.537 & 0.604 & 0.627 & 0.598 & 0.592 \\
& & & 1 & 0.530 & 0.602 & 0.621 & 0.628 & 0.595 & 0.543 & 0.574 & 0.619 & 0.602 & 0.585 \\
& & & 2 & 0.532 & 0.616 & 0.591 & 0.625 & 0.591 & 0.542 & 0.579 & 0.584 & 0.585 & 0.573 \\
& & & 3 & 0.532 & 0.612 & 0.562 & 0.620 & 0.581 & 0.531 & \textbf{0.589} & 0.577 & 0.619 & 0.579 \\
& & & 4 & 0.508 & 0.625 & 0.529 & 0.592 & 0.564 & \textbf{0.549} & \textbf{0.562} & 0.564 & 0.586 & 0.565 \\

\rowcolor{cyan!10}
& & & 0 & 0.875 & \textbf{0.985} & \textbf{0.914} & \textbf{0.967} & \textbf{0.935} & 0.875 & \textbf{0.985} & \textbf{0.914} & \textbf{0.967} & \textbf{0.935} \\
\rowcolor{cyan!10}
& & & 1 & 0.761 & \textbf{0.907} & 0.830 & \textbf{0.862} & \textbf{0.840} & 0.702 & \textbf{0.738} & \textbf{0.749} & \textbf{0.780} & \textbf{0.742} \\
\rowcolor{cyan!10}
& & & 2 & 0.742 & \textbf{0.857} & \textbf{0.799} & \textbf{0.824} & \textbf{0.806} & \textbf{0.648} & \textbf{0.605} & \textbf{0.649} & \textbf{0.703} & \textbf{0.651} \\
\rowcolor{cyan!10}
& & & 3 & 0.684 & \textbf{0.810} & \textbf{0.759} & \textbf{0.806} & \textbf{0.765} & \textbf{0.577} & 0.551 & \textbf{0.581} & \textbf{0.665} & \textbf{0.594} \\
\rowcolor{cyan!10}
\multirow{-5}{*}{\cellcolor{cyan!10}\textbf{Face-D$^2$CL}} &
\multirow{-5}{*}{\cellcolor{cyan!10}-} &
\multirow{-5}{*}{\cellcolor{cyan!10}$\checkmark$} &
4 & \textbf{0.639} & \textbf{0.741} & \textbf{0.718} & \textbf{0.756} & \textbf{0.714} & 0.541 & 0.486 & \textbf{0.591} & \textbf{0.675} & \textbf{0.573} \\
\bottomrule
\end{tabular}
}

\smallskip
\centering
\captionof{table}{Robustness results under Gaussian noise and median blur perturbations. Each perturbation is applied at five intensity levels (Level 0: unperturbed; Levels 1--4: increasing strength). Results are reported in AUC.}
\label{tab:robustness_noise_blur}
\tiny
\setlength{\tabcolsep}{1.4pt}
\renewcommand{\arraystretch}{1.00}
\resizebox{\textwidth}{!}{%
\begin{tabular}{c c c c c c c c c c c c c c}
\toprule
\multirow{2}{*}{Method} &
\multirow{2}{*}{Venue} &
\multirow{2}{*}{No-Replay} &
\multirow{2}{*}{Level} &
\multicolumn{5}{c}{Gaussian Noise} &
\multicolumn{5}{c}{Median Blur} \\
\cmidrule(lr){5-9} \cmidrule(lr){10-14}
& & & & FF++ & DFDCP & DFD & CDF2 & Avg & FF++ & DFDCP & DFD & CDF2 & Avg \\
\midrule
\multirow{5}{*}{DFIL} &
\multirow{5}{*}{ACM MM'23} &
\multirow{5}{*}{$\times$} &
0 & \textbf{0.962} & 0.884 & 0.888 & 0.922 & 0.914 & \textbf{0.962} & 0.884 & 0.888 & 0.922 & 0.914 \\
& & & 1 & \textbf{0.875} & 0.757 & 0.864 & 0.830 & 0.832 & \textbf{0.944} & 0.888 & 0.889 & 0.933 & 0.914 \\
& & & 2 & 0.721 & 0.670 & 0.807 & 0.686 & 0.721 & \textbf{0.896} & 0.865 & 0.817 & 0.878 & 0.864 \\
& & & 3 & 0.657 & 0.649 & 0.668 & 0.590 & 0.641 & \textbf{0.845} & 0.814 & 0.752 & 0.822 & 0.808 \\
& & & 4 & 0.580 & 0.555 & 0.564 & 0.533 & 0.558 & \textbf{0.785} & 0.760 & 0.693 & 0.770 & 0.752 \\

\multirow{5}{*}{SUR-LID} &
\multirow{5}{*}{CVPR'25} &
\multirow{5}{*}{$\times$} &
0 & 0.537 & 0.899 & 0.410 & 0.409 & 0.564 & 0.537 & 0.899 & 0.410 & 0.409 & 0.564 \\
& & & 1 & 0.539 & 0.857 & 0.419 & 0.431 & 0.562 & 0.548 & 0.907 & 0.393 & 0.460 & 0.577 \\
& & & 2 & 0.518 & 0.777 & 0.482 & 0.560 & 0.584 & 0.537 & 0.915 & 0.382 & 0.525 & 0.590 \\
& & & 3 & 0.475 & 0.673 & 0.499 & 0.583 & 0.558 & 0.523 & 0.875 & 0.404 & 0.546 & 0.587 \\
& & & 4 & 0.510 & 0.578 & 0.503 & 0.574 & 0.541 & 0.505 & 0.811 & 0.443 & 0.570 & 0.582 \\

\multirow{5}{*}{SAIDO} &
\multirow{5}{*}{CVPR'26} &
\multirow{5}{*}{$\checkmark$} &
0 & 0.537 & 0.604 & 0.627 & 0.598 & 0.592 & 0.537 & 0.604 & 0.627 & 0.598 & 0.592 \\
& & & 1 & 0.542 & 0.631 & 0.632 & 0.645 & 0.613 & 0.519 & 0.578 & 0.596 & 0.567 & 0.565 \\
& & & 2 & 0.566 & 0.651 & 0.611 & 0.694 & 0.631 & 0.505 & 0.551 & 0.556 & 0.538 & 0.538 \\
& & & 3 & 0.505 & 0.424 & 0.538 & 0.418 & 0.471 & 0.501 & 0.519 & 0.537 & 0.518 & 0.519 \\
& & & 4 & 0.525 & 0.467 & 0.512 & 0.430 & 0.484 & 0.502 & 0.512 & 0.531 & 0.521 & 0.517 \\

\rowcolor{cyan!10}
& & & 0 & 0.875 & \textbf{0.985} & \textbf{0.914} & \textbf{0.967} & \textbf{0.935} & 0.875 & \textbf{0.985} & \textbf{0.914} & \textbf{0.967} & \textbf{0.935} \\
\rowcolor{cyan!10}
& & & 1 & 0.861 & \textbf{0.938} & \textbf{0.930} & \textbf{0.949} & \textbf{0.920} & 0.852 & \textbf{0.981} & \textbf{0.896} & \textbf{0.963} & \textbf{0.923} \\
\rowcolor{cyan!10}
& & & 2 & \textbf{0.841} & \textbf{0.925} & \textbf{0.912} & \textbf{0.921} & \textbf{0.900} & 0.824 & \textbf{0.971} & \textbf{0.881} & \textbf{0.946} & \textbf{0.906} \\
\rowcolor{cyan!10}
& & & 3 & \textbf{0.801} & \textbf{0.875} & \textbf{0.852} & \textbf{0.841} & \textbf{0.842} & 0.794 & \textbf{0.945} & \textbf{0.829} & \textbf{0.905} & \textbf{0.868} \\
\rowcolor{cyan!10}
\multirow{-5}{*}{\cellcolor{cyan!10}\textbf{Face-D$^2$CL}} &
\multirow{-5}{*}{\cellcolor{cyan!10}-} &
\multirow{-5}{*}{\cellcolor{cyan!10}$\checkmark$} &
4 & \textbf{0.759} & \textbf{0.765} & \textbf{0.710} & \textbf{0.718} & \textbf{0.738} & 0.741 & \textbf{0.902} & \textbf{0.768} & \textbf{0.858} & \textbf{0.817} \\
\bottomrule
\end{tabular}
}

\end{minipage}
}]

\clearpage
\twocolumn[{%
\begin{minipage}{\textwidth}
\centering
\captionof{table}{Training overhead of RF-EWC and D-OGC.}
\label{tab:ewc_ogc_overhead}
\scriptsize
\setlength{\tabcolsep}{4pt}
\renewcommand{\arraystretch}{0.88}
\resizebox{0.72\textwidth}{!}{%
\begin{tabular}{@{}lccc@{}}
\toprule
Setting & Time / epoch (s) & Peak Mem (GB) & GFLOPs \\
\midrule
\multicolumn{4}{l}{\textit{RF-EWC vs.\ standard Fisher}} \\
w/ RF-EWC (main) & 117.44 & 20.98 & 24042 \\
Standard Fisher & 117.39 & 20.98 & 24042 \\
Delta & +0.05 & 0 & 0 \\
\midrule
\multicolumn{4}{l}{\textit{D-OGC vs.\ w/o D-OGC}} \\
w/ D-OGC (main) & 38.61 & 26.35 & 233.88 \\
w/o D-OGC & 35.78 & 26.30 & 233.88 \\
Delta & +2.83 & +0.05 & 0 \\
\bottomrule
\end{tabular}
}

\medskip
\centering
\captionof{table}{Real-face manifold preservation on Protocol~2 ($k$-NN subset pseudo-separation, $k{=}15$; lower is better).}
\label{tab:manifold}
\large
\setlength{\tabcolsep}{5pt}
\renewcommand{\arraystretch}{1.35}
\begin{tabular}{@{}lc@{}}
\toprule
Method & Subset pseudo-separation \\
\midrule
DFFreq & 0.288 \\
SUR-LID & 0.309 \\
SAIDO & 0.366 \\
\rowcolor{cyan!10}
\textbf{Face-D\(^2\)CL} & \textbf{0.279} \\
\bottomrule
\end{tabular}

\medskip
\centering
\captionof{table}{Eight-task continual learning after Task~8 (ACC). Tasks 5--8: SimSwap, FOMM, SiT-XL/2, CollabDiff.}
\label{tab:eight_task}
\small
\setlength{\tabcolsep}{2.5pt}
\renewcommand{\arraystretch}{0.95}
\resizebox{\textwidth}{!}{%
\begin{tabular}{@{}l *{8}{c} c@{}}
\toprule
Method & FF++ & DFDCP & DFD & CDF2 & SimSwap & FOMM & SiT-XL/2 & CollabDiff & AA / AF \\
\midrule
SAIDO & 98.1 & 85.2 & 98.7 & 84.8 & 96.8 & 96.2 & 94.4 & 96.2 & 93.8 / 2.6 \\
SUR-LID & 86.7 & 81.6 & 85.4 & 82.2 & 98.8 & 97.8 & 93.6 & 95.4 & 90.2 / 7.5 \\
\rowcolor{cyan!10}
\textbf{Face-D\(^2\)CL} & 99.7 & 89.2 & 97.4 & 94.2 & 97.6 & 94.2 & 89.8 & 94.6 & \textbf{94.6} / \textbf{2.5} \\
\bottomrule
\end{tabular}
}

\medskip
\centering
\captionof{table}{Nine-task AIGC continual learning after Task~9 (ACC).}
\label{tab:aigc9}
\small
\setlength{\tabcolsep}{3pt}
\renewcommand{\arraystretch}{0.95}
\resizebox{\textwidth}{!}{%
\begin{tabular}{@{}l *{9}{c} c@{}}
\toprule
Method & ADM & GLIDE & ProGAN & SAGAN & BigGAN & wukong & SD1.5 & Midj. & VQDM & AA / AF \\
\midrule
SAIDO & 95.2 & 99.2 & 96.9 & 99.9 & 87.5 & 90.7 & 97.0 & 69.8 & 99.9 & 92.9 / 7.1 \\
\rowcolor{cyan!10}
\textbf{Face-D\(^2\)CL} & 99.3 & 98.7 & 97.1 & 98.1 & 96.2 & 97.5 & 90.5 & 86.7 & 98.9 & \textbf{95.9} / \textbf{2.4} \\
\bottomrule
\end{tabular}
}

\vspace{1.2\baselineskip}
\end{minipage}
}]

\clearpage
\section{Mechanism Analyses}
\label{sec:mechanism}

\subsection{Cross-branch CKA Analysis}
\label{sec:cka}

\begin{figure}[t]
\centering
\resizebox{0.90\columnwidth}{!}{\includegraphics{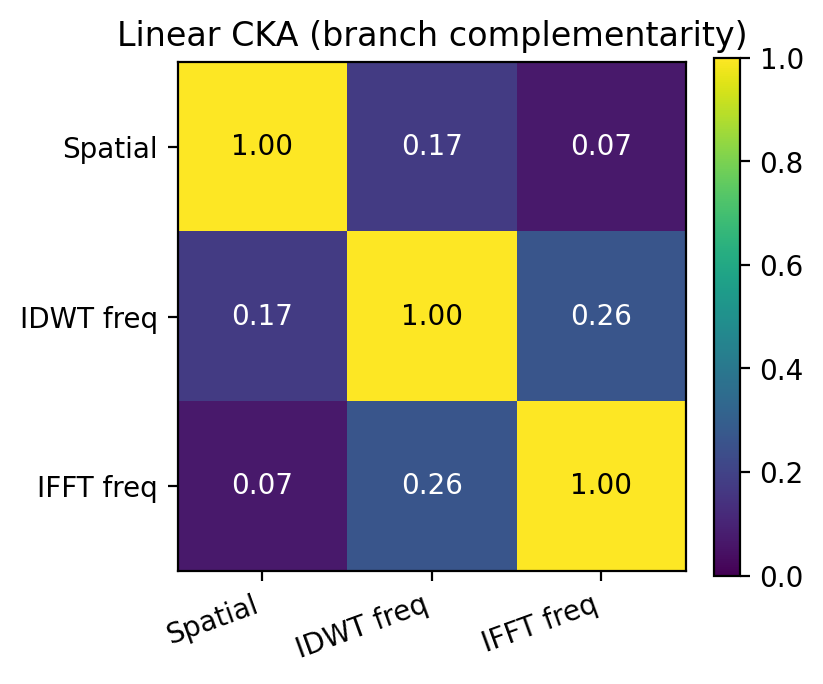}}
\caption{Cross-branch CKA among Spatial, Wavelet (IDWT), and Fourier (IFFT) features (centered linear CKA; Spatial--Wavelet $0.17$, Spatial--Fourier $0.07$, Wavelet--Fourier $0.26$).}
\label{fig:cka}
\end{figure}

Branch ablation shows that each domain helps accuracy, but not whether the three experts encode overlapping or complementary subspaces. To check redundancy, we compute linear Centered Kernel Alignment (CKA) on centered Spatial, Wavelet (IDWT), and Fourier (IFFT) features. As shown in Figure~\ref{fig:cka}, diagonal entries are $1.00$ by construction, while off-diagonal scores remain low (Spatial--Wavelet $0.17$, Spatial--Fourier $0.07$, Wavelet--Fourier $0.26$). Together with the AA drops in Table~\ref{tab:ablation_multidomain}, this supports that the three experts capture complementary forgery cues rather than duplicated views, which justifies retaining all three branches and domain-wise D-OGC updates.

\subsection{D-OGC Algorithm}
\label{sec:dogc_algorithm}
\begin{center}
\footnotesize
\begin{tabular}{@{}p{0.94\columnwidth}@{}}
\toprule
\textbf{Algorithm: Domain-wise D-OGC} \\
\midrule
\textbf{Initialize} $S_d\gets0$, $N_d\gets0$, $\mathcal H_d\gets[\,]$, and $\mathcal C_d\gets\varnothing$ for each $d\in\{S,W,F\}$. \\
\textbf{for} each task $k$ \textbf{do} \\
\quad \textbf{for} each minibatch $b$ \textbf{do} compute unprojected gradients $\{g_{k,b,d}\}_{d\in\{S,W,F\}}$. \\
\quad\quad \textbf{for} each expert $d\in\{S,W,F\}$ \textbf{do} set $S_d\gets S_d+g_{k,b,d}$ and $N_d\gets N_d+1$. \\
\quad\quad\quad \textbf{if} $\mathcal H_d\neq[\,]$ and $\mathcal C_d=\varnothing$, set $r_d\gets |\mathcal H_d|^{-1}\sum_{h\in\mathcal H_d}h$; if $\|r_d\|_2>\epsilon$, set $\mathcal C_d\gets r_d/\|r_d\|_2$. \\
\quad\quad\quad \textbf{if} $\mathcal C_d\neq\varnothing$, set $\tilde g_{k,b,d}\gets g_{k,b,d}-(g_{k,b,d}^{\top}\mathcal C_d)\mathcal C_d$; otherwise set $\tilde g_{k,b,d}\gets g_{k,b,d}$. \\
\quad\quad \textbf{end for}; update the experts using $\{\tilde g_{k,b,d}\}_{d\in\{S,W,F\}}$. \\
\quad \textbf{end for} \\
\quad \textbf{for} each $d$ \textbf{do} set $\bar g_{k,d}\gets S_d/\max(1,N_d)$; if $\|\bar g_{k,d}\|_2>\epsilon$, append $h_{k,d}\gets\bar g_{k,d}/\|\bar g_{k,d}\|_2$ to $\mathcal H_d$; reset $S_d,N_d\gets0$ and $\mathcal C_d\gets\varnothing$. \\
\textbf{end for} \\
\bottomrule
\end{tabular}
\end{center}

\subsection{D-OGC Numerical Stability in LoRA Space}
Orthogonal projection in D-OGC removes the component of the current LoRA gradient along the historical cache \(g_{\text{hist}}\), which raises a natural concern: the constraint may overly shrink updates and harm plasticity to new forgery types. Under Protocol~1 we record after each backward pass
\[
\|g_{\mathrm{post}}\|/\|g_{\mathrm{pre}}\|,
\]
where \(g_{\mathrm{pre}}\) is the LoRA gradient before projection and \(g_{\mathrm{post}}=\tilde{g}\) is the gradient after the projection in Eq.~\eqref{eq:dogc}. Ratios closer to \(1\) indicate weaker attenuation of the update. On the first task (FF++), \(g_{\text{hist}}\) is empty, so the ratio is identically \(1.000\). On the incremental tasks the mean ratios are \(0.792\) (DFDCP), \(0.766\) (DFD), and \(0.744\) (CDF2), with step-wise ranges in \([0.58,0.91]\) (Table~\ref{tab:dogc_norm_ratio}). Thus every stage retains at least about three quarters of the pre-projection gradient energy. Together with the \(2.9\%\) AA drop when D-OGC is removed (Table~\ref{tab:ablation_dualcl}), these results indicate that domain-wise projection improves detection accuracy without collapsing the effective update needed to adapt domain experts.

\begin{center}
\captionof{table}{D-OGC gradient-norm ratio $\|g_{\mathrm{post}}\|/\|g_{\mathrm{pre}}\|$ on Protocol~1 (mean over training steps; brackets: $[\min,\max]$).}
\label{tab:dogc_norm_ratio}
\small
\setlength{\tabcolsep}{4pt}
\renewcommand{\arraystretch}{0.90}
\begin{tabular}{@{}lcccc@{}}
\toprule
Task & FF++ & DFDCP & DFD & CDF2 \\
\midrule
Mean ratio & 1.000 & 0.792 & 0.766 & 0.744 \\
Range & $[1.00,1.00]$ & $[0.67,0.90]$ & $[0.59,0.91]$ & $[0.58,0.91]$ \\
\bottomrule
\end{tabular}
\end{center}

\section{Backbone Selection}
To check whether Face-D\(^2\)CL depends on a specific CLIP encoder, we swap the backbone on Protocol~1 while keeping the multi-domain experts and dual continual learning modules. As shown in Table~\ref{tab:backbone}, CLIP-ViT-L/14 achieves the highest AA and lowest AF; ResNet-50 / 101 / 152 all obtain AA no lower than 0.97, and DINOv2-B/14 no lower than 0.95. Within ResNet, deeper models slightly improve AA but keep clearly higher AF than CLIP; DINOv2-B/14 beats DINOv2-S/14 by 2.7\% AA yet still trails CLIP. We therefore adopt CLIP-ViT-L/14 as the default backbone, while the remaining encoders confirm that the proposed constraints remain beneficial across feature extractors.

\begin{table}[t]
\centering
\caption{Backbone comparison on Protocol~1 (ACC).}
\label{tab:backbone}
\large
\setlength{\tabcolsep}{5pt}
\renewcommand{\arraystretch}{1.30}
\begin{tabular}{@{}lcc@{}}
\toprule
Encoder & AA & AF \\
\midrule
CLIP-ViT-L/14 & 0.985 & 0.006 \\
ResNet-50 & 0.971 & 0.093 \\
ResNet-101 & 0.975 & 0.122 \\
ResNet-152 & 0.977 & 0.106 \\
DINOv2-S/14 & 0.925 & 0.019 \\
DINOv2-B/14 & 0.952 & 0.041 \\
\bottomrule
\end{tabular}
\end{table}

\end{document}